%% file: main.tex
\documentclass{article}
\usepackage[utf8]{inputenc}
\usepackage{authblk}
\usepackage{setspace}
\usepackage[margin=1.25in]{geometry}
\usepackage{graphicx}

\graphicspath{ {./figures/} }
\usepackage{subcaption}
\usepackage{amsmath}
\usepackage{amssymb}

\usepackage[style=ieee, 
citestyle=numeric-comp,
sorting=none]{biblatex}
\usepackage{times}
\usepackage{multirow}
\usepackage{booktabs}
\usepackage{tabularx}
\usepackage{relsize}
\usepackage{caption}
\usepackage{subfiles}

\title{Deep Multi-view Image Fusion for Soybean Yield Estimation in Breeding Applications}

\author[1$\dag$]{Luis G Riera}
\author[2$\dag$]{Matthew E. Carroll}
\author[1]{Zhisheng Zhang}

\author[2]{Johnathon M. Shook}
\author[1]{Sambuddha Ghosal}
\author[3]{Tianshuang Gao}

\author[2]{Arti Singh}

\author[1]{Sourabh Bhattacharya}
\author[1]{Baskar Ganapathysubramanian}
\author[*2]{Asheesh K. Singh}
\author[*1]{Soumik Sarkar}

\affil[1]{Department of Mechanical Engineering, Iowa State University, Ames, Iowa, USA}
\affil[2]{Department of Agronomy, Iowa State University, Ames, Iowa, USA}
\affil[3]{Department of Computer Science, Iowa State University, Ames, Iowa, USA}
\affil[*]{Corresponding authors. Email: singhak@iastate.edu \& soumiks@iastate.edu}
\affil[$\dag$]{These authors contributed equally to this work.}

\date{}

\begin{document}

\maketitle

\begin{abstract}
Reliable seed yield estimation is an indispensable step in plant breeding programs geared towards cultivar development in major row crops. The objective of this study is to develop a machine learning (ML) approach adept at soybean [\textit{Glycine max} L. (Merr.)] pod counting to enable genotype seed yield rank prediction from in-field video data collected by a ground robot. To meet this goal, we developed a multi-view image-based yield estimation framework utilizing deep learning architectures. Plant images captured from different angles were fused to estimate the yield and subsequently to rank soybean genotypes for application in breeding decisions. We used data from controlled imaging environment in field, as well as from plant breeding test plots in field to demonstrate the efficacy of our framework via comparing performance with manual pod counting and yield estimation. 
Our results demonstrate  the promise of ML models in making breeding decisions with significant reduction of time and human effort, and opening new breeding methods avenues to develop cultivars. 
\end{abstract}


\section{Introduction}
\label{sec:Introduction}
\medskip
\subfile{Introduction}

\section{Materials and Methods}
\label{sec:Methods}
\medskip
\subfile{Methods}

\section{Results and Discussion}
\label{sec:Discussion}
\medskip
\subfile{Discussion}

\section{Conclusion and Future work}
\label{sec:conclusion}
\medskip
\subfile{conclusion}

\break
\section*{Supplementary Material}
\label{sec:Supplementary}
\medskip
\subfile{Supplementary}

\pagebreak

\printbibliography

\end{document}

%% file: Introduction.tex

Plant breeding programs worldwide rely on yield testing to make selections and advancement decisions towards the development of new varieties. A vital component in this process is the growing and harvesting of an inordinate number of plots at several locations each year, incurring substantial costs and resource allocations burdening the economics of a breeding program. The need to assess tens of thousands of genotypes in a program is necessitated by the inherent requirement to work with a desired level of and expand the genetic variance for a higher response to selection \cite{hazel1942efficiency}. Therefore, the need to accurately measure or predict yield has motivated researchers to constantly develop modern tools in genomics \cite{xavier2016assessing}, \cite{zhang2016genome} and phenomics \cite{harfouche2019accelerating}, \cite{rebetzke2019high}, \cite{zhang2019crop} . 

One of the avenues to yield prediction is through the fusion of high dimensional phenotypic trait data using machine learning (ML) approaches to provide plant breeders the tools to do in-season seed yield (SY) prediction~\cite{parmley2019machine}, and fusing ML and optimization techniques to identify a suite of in-season phenotypic traits collected from multiple sensors that decrease the dependence on resource-intensive end-season phenotyping in breeding programs \cite{parmley2019development}. Other avenues have been through integrating weather and genetic information in conjunction with deep time series attention models for crop SY prediction~\cite{jiang2020predicting},~\cite{gangopadhyayyieldneurips2019}. 

These advances in ML methods and earnest effort to collect large data sets is commendable, and has a positive role in numerous scenarios; however, these approaches do not work for plant breeding programs of all sizes, geographical regions and crops. One less explored approach is using simple camera (tri-band digital) to estimate plant reproductive organs and estimate SY. If imaging is coupled with automated ground robotic systems, breeders can compute plot SY to make breeding decisions in an efficient manner. Gao et al.~\cite{gao2018novel} deployed a low cost lightweight distributed multiple robot system for soybean phenotypic data collection, which demonstrates a usable platform to meet yield estimation requirements. We envision that an automated data collection platform (i.e., ground robot) with sensors (i.e., digital camera) creates a framework to estimate SY in field conditions from breeding plots where genotypes are assessed for their merit and commercialization potential. The motivation for this challenge is to provide a timely and cost effective solution for SY estimation in field plots. This framework can be deployed in a breeding pipeline to improve the capability to obtain high quality SY data without the need to machine harvest all plots, one of the most time and resource intensive steps in plant breeding. To provide a context of the time and resource investment, ~50-100 soybean yield plots can be combined in one hour at each testing field site depending on plot sizes etc, by a 1-2 person work crew. Plant breeding programs are also at the mercy of weather events. For example, excessive rainfall in fall season in North America soybean growing regions often complicates machine harvest, bringing the entire program to a halt causing significant delay in data analysis and advancement decisions for winter nursery operations or for the next season. Also, breeding decisions for making selections and advancements need to wait for machine harvest data delaying breeding cycles and turnaround times. 

We suggest a balanced and strategic approach by using small weight autonomous ground robots. This is applicable to early generation testing, progeny row stage, and also for preliminary and unreplicated advanced replicated yield trial stage, where genotypes can be grown in multiple locations but only one location needs to be harvested to obtain seed source for next season planting. Additionally, in replicated tests, only one replication will be harvested for seed source. Furthermore, such a breeding pipeline will empower breeding programs to operate in wet soil conditions where machine harvest is not possible. In all these above scenarios, except the harvest plot, all other plots will be imaged using the ground robot and yield estimated using the application of ML models by obtaining information on plant reproductive organs of economic importance, such as SY. These situations and the integration of yield estimation in breeding methods and pipelines using ground robots, can remove the need to harvest all plots from all locations, to save time and resources as described earlier. 

Computer vision models for crop yield estimation have been proposed in the past. However, such models are primarily built for larger fruit trees, with potentially less background clutter and occlusion compared to soybean plants and pods~\cite{fernandez2014multisensory},~\cite{Nuske2011},~\cite{bulanon2009image},~\cite{gongal2016apple}, and~\cite{linker2018machine}. Recently proposed TasselNet~\cite{lu2017tasselnet} attempts to count maize tassels based on a single aerial image.

In this context, we develop a deep learning framework for soybean pod detection to predict the number of pods in each plot using multiple views of plants taken by simple RGB cameras. This information is then used as a proxy to estimate seed yield. We also test the usefulness of this model on images collected by a ground robot capable of taking RGB images that are processed by the deep learning model. While the proposed pod counting and yield estimation approach can be used in an offline manner using single or multi-view images of plots, we further develop a plant detector and tracker (from soybean plots), enabling our approach to also be used in an online manner by a ground robot using on-board processors and edge computing. The rest of the paper is organized as follows. Section 2 (Materials and Methods) presents the data collection and processing steps along with the ML framework for pod counting and plant detection/tracking from plots. In Section 3 (Results and Discussion), we present the performance of our proposed framework and discuss the feasibility and promise of our approach in breeding programs by comparing performance with manual ground truth. Finally, the paper is summarized and concluded with directions of future work.

%% file: Methods.tex

In this section, we describe data collection and pre-processing steps essential for this project, followed by details on the deep learning framework for pod detection and counting, a framework for plant detection and tracking from soybean plots that can enable real-time pod counting (and yield estimation) using a ground robot with on-board processing.


\subsection{Data acquisition and pre-processing}
\label{ssec:ExperimentalSetup}
In this paper, we used two different data sets for ML model development as well as validation. The first one is a \textit{control data set} acquired in an outdoor (i.e., field) environment with an effort to use optimal lighting and other imaging environmental settings. The second \textit{in-field data set} is a more realistic one, collected in the field with soybean crops with diverse environmental variability. Details of these data sets are provided below along with the specific pre-processing steps used in this study. 

\begin{figure}
	\centering
	\begin{subfigure}{.3\textwidth}
		\centering
		\includegraphics[width = 4.25cm, height = 5cm]{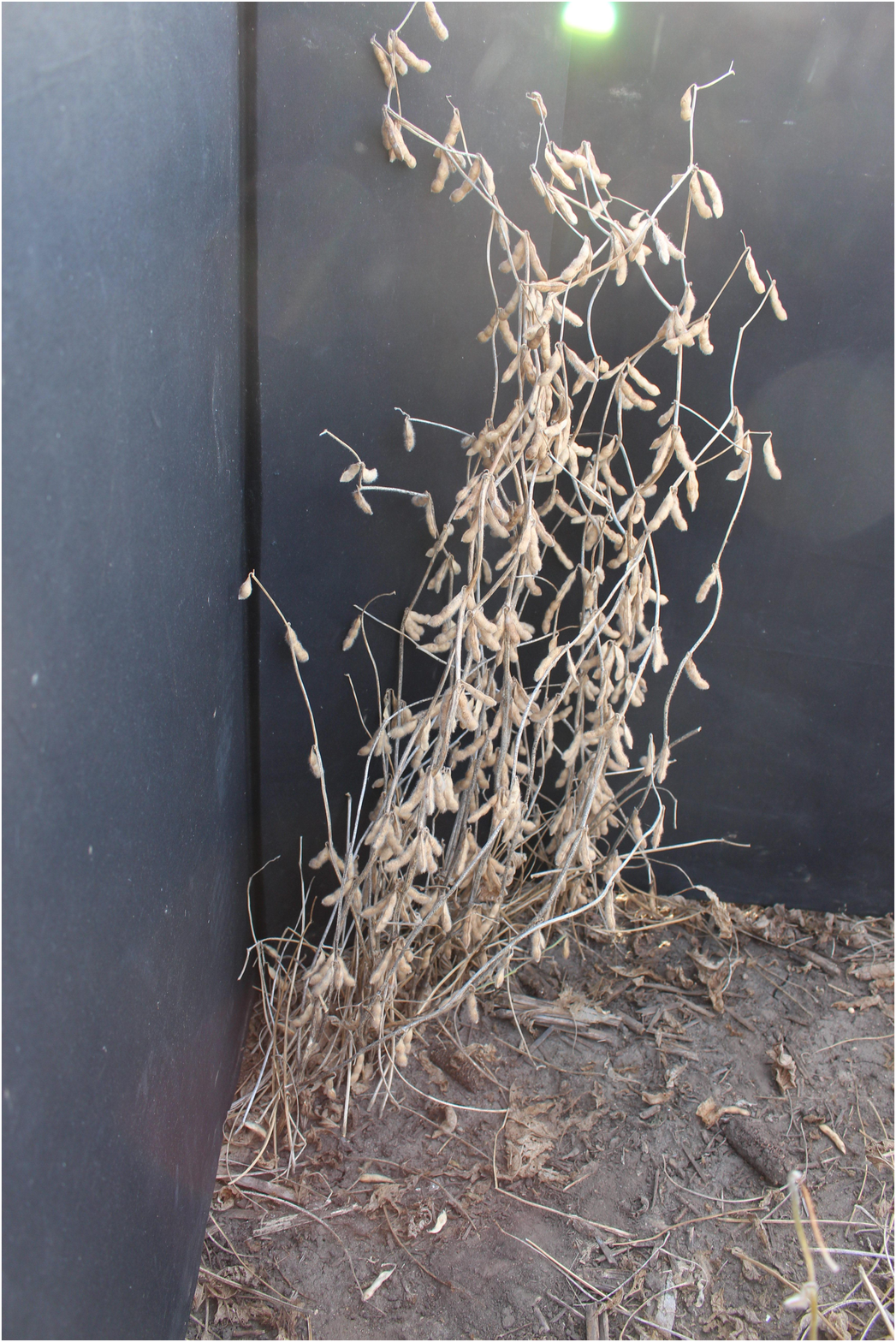}
		\caption{Left view}
		\label{fig:BlakBackgroundLeftView}
	\end{subfigure}%
	\begin{subfigure}{.3\textwidth}
		\centering
		\includegraphics[width = 4.25cm, height = 5cm]{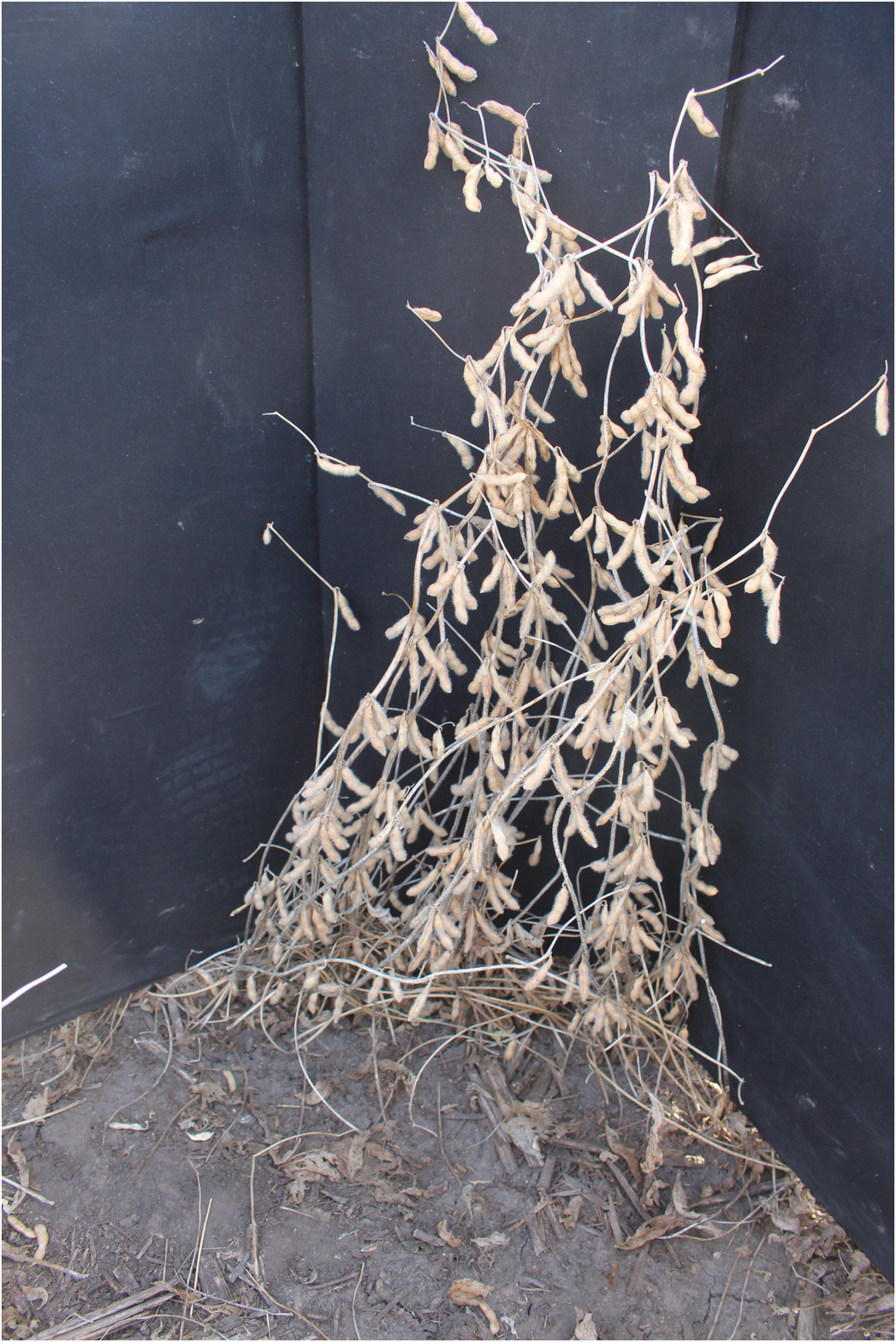}
		\caption{Front view}
		\label{fig:BlakBackgroundFrontView}
	\end{subfigure}%
	\begin{subfigure}{.3\textwidth}
		\centering
		\includegraphics[width = 4.25cm, height = 5cm]{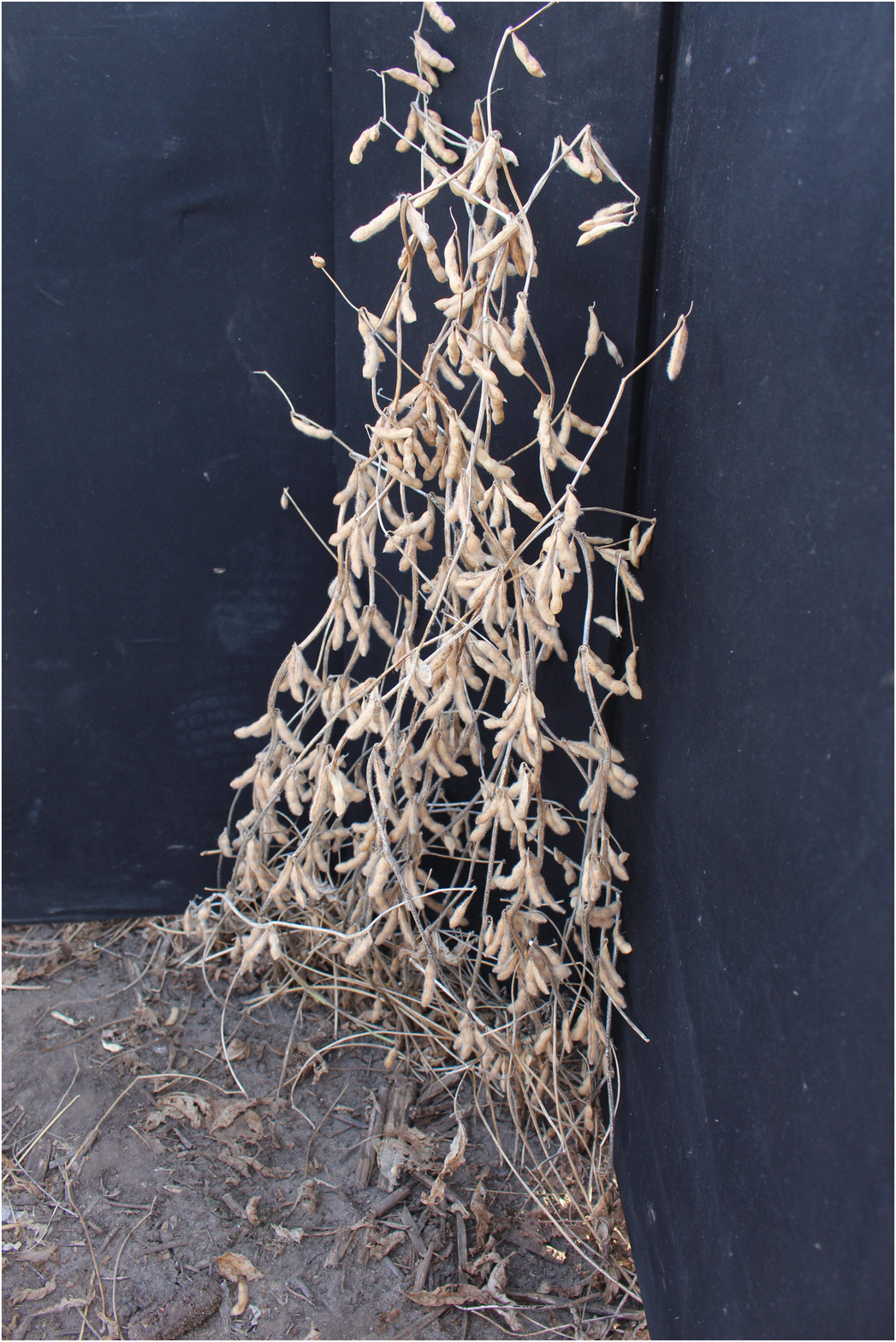}
		\caption{Right view}
		\label{fig:BlakBackgroundRightView}
	\end{subfigure}
	\caption{Sample images from the \textit{control data set} along the background trifold used to remove background noise. These three views correspond to the same plot with multiple plants.}
	\label{fig:BlackBackgroundTrifold}
	
\end{figure}

\textbf{Control Data Set:} The images in this set were collected from random 30.5cm sub-sections with soybean plants from matured soybean plots in 2014 (145 sub-sections) and 2015 (154 sub-sections). Three images were taken for each sub-section using a tri-fold black background, which was used to remove background artifacts from the images as seen in Fig.~\ref{fig:BlackBackgroundTrifold}. Images were taken with a Canon EOS Rebel T5 in the RAW 18 mega pixel format, and were converted to jpg for processing. The focus and white balance were set to auto, and were adjusted as per the prevailing conditions. Upon the completion of imaging, plants from the 30.5cm sections were cut from the ground level using a sharp clipper, and bundled in a bag for pod counting. Care was taken to ensure no plant part loss occurred, enabling accurate pod counting. One sub-section in 2015 had four images that were taken instead of three images, so we did not use information from this sub-section. 

\begin{figure}
	\centering
	\begin{subfigure}{.48\textwidth}
		\centering
		\includegraphics[width =6.9cm, height = 5.75cm]
		{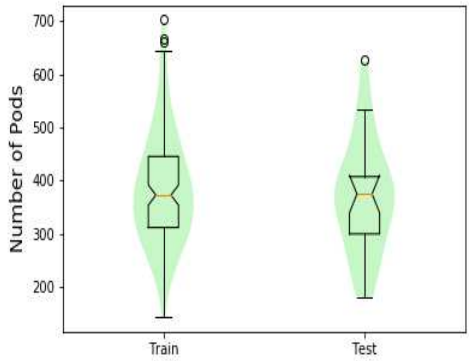}
		\caption{Pod count distribution for training and test \\ subsets within the \textit{control data set}}
		\label{fig:ControlViolinPlot}
	\end{subfigure}
	\begin{subfigure}{.48\textwidth}
		\centering
		\includegraphics[width = 6.9cm, height = 5.75 cm]
		{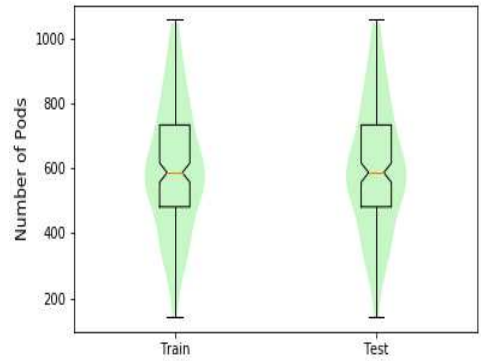}
		\caption{Pod count distribution for training and test \\ subsets within the \textit{in-field data set}}
		\label{fig:In-fieldViolinPlot}
	\end{subfigure}
	
	\caption{Pod count distributions for control and in-field data sets}
	\label{fig:DatasetsDitribution}
\end{figure}

Expert raters labeled images for the 298 sub-sections using the VIA (VGG Image Annotator) image labeling software~\cite{dutta2018vgg} to create the bounding boxes for pods. This data was split into two subsets for training (247 plots) and testing (51 plots) the ML models. Statistical characteristics of these data sets are provided in Table~\ref{tab:Dataset} and in Fig.~\ref{fig:ControlViolinPlot}.

\begin{figure}
	\begin{subfigure}{.5\textwidth}
		\centering
		\includegraphics[width = 5.2cm, height = 5.9cm]{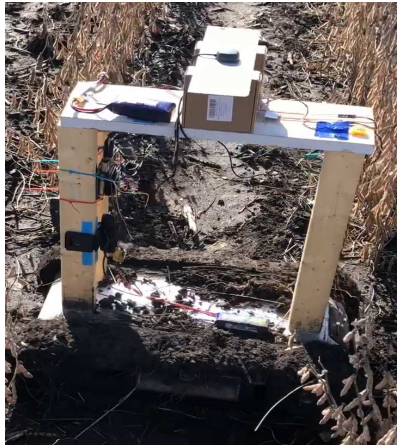}
		\caption{Sensor setup on the ground robot}
		\label{fig:BotPhoto}
	\end{subfigure}
	\begin{subfigure}{.5\textwidth}
		\centering
		\includegraphics[width = 7.5cm, height = 5. cm]{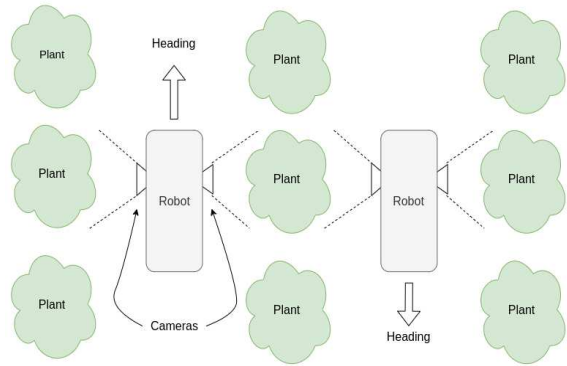}
		
		\caption{Diagram demonstrating how the ground robot typically traverse a  soybean field test}
		\label{fig:BotCameraSetup}
	\end{subfigure}%

    \caption{The top row illustrate the ground robots' sensors setup and how it traverse the field, while images from (c) to (h) in bottom two rows are typical video frames images, taken by the ground robot, viewing the same plot from the North and South sides.}
\end{figure}

\textbf{In-field Data Set:} This data set consisted of images taken from a ground robot \cite{robotsinc}. The ground robot was outfitted with a wooden frame with mounted webcam (Logitech C920) to capture images, ensuring that full length of the soybean plants in each plot were imaged in each frame (Fig .\ref{fig:BotPhoto}). Videos were captured at a frame rate of 30 frames per second and at 720p resolution. The camera configuration permitted filming the two sides of the soybean plots with fewer passes (Fig .\ref{fig:BotCameraSetup}). 
This data was collected from the USDA GRIN mini core collection \cite{oliveira2010establishing} genotypes grown in a field near Ames, IA, in 15.24cm length and 76.2cm plot to plot distance. All plots were hand harvested at the R8 growth stage \cite{fehr1977stages}, after they were imaged with the ground robot. The plant height of genotypes from these plots ranged from 25cm to 108cm with a median height of 70cm and a standard deviation of 15.99 cm. All plots were also rated for lodging on a scale of 1-5 with a score of 1 being all plants being erect, and a score of 5 being prostrate. In this study, 46\% of the plots were scored as a 1, 24\% were scored as a 2, 13\% were scored as a 3, 10\% were scored as a 4, and 7\% were scored as a 5. The genotypes varied in pod and pubescence colors, with 60\% of the genotypes having a brown pod color and 40\% had a tan pod color. 26\% of the lines had light tawny pubescence, 32\% had tawny pubescence, and 42\% had gray pubescence. Genotypes could be further separated into elite, diverse and PI types representing commercial varieties to unimproved introductions \cite{song2017genetic}. The overall mean of the elite genotypes was 641 pods per plot (range of 313 to 1038 pods), diverse genotypes had a mean of 623 (range of 142 to 1058 pods), and PI had a mean of 466 pods (range of 150 to 805 pods).

Overall, we selected 124 plots in this data set. An expert rater determined the start and end of the frame sequences for each plot in each pass. This is to ensure that the frames accurately correspond to the plots for which manual pod counting was performed to obtain ground truth. The number of video frames per plot ranged from 11 to 98, with a median of 38 frames per plot. As we will consider only a few frames (1 to 3 in this study) per side of a plot to estimate yield, different sets of frames corresponding to a plot can be taken as different samples. We use this logic to perform data augmentation and form training and test sets with 167 and 43 plot samples, respectively (i.e., total of 210 samples). Statistical characteristics of these data sets are provided in Table~\ref{tab:Dataset} and in Fig.~\ref{fig:In-fieldViolinPlot}.

    \begin{table}[t]
    \small
   	\renewcommand{\arraystretch}{1.4}
   	\caption{Descriptive statistics of the datasets include the control and in-field data sets}
   	\label{tab:Dataset}
   	\centering
   	
   	\makebox[\textwidth][c]{
   		\begin{tabular}{c l c c | c c }
   			\toprule
   	        \multicolumn{2}{c}{\multirow{2}{*}{Datasets}}	& \multicolumn{2}{c}{\textbf{Control set}} & \multicolumn{2}{|c}{\textbf{In-field set}} \\
   			\cline{3-6}
   			&	\multicolumn{1}{c}{} & \textit{Train} & \textit{Test} & \textit{Train} & \textit{Test} \\
   			\hline
   			\multicolumn{2}{c}{No. Plots Annotated}  & \textit{247} & \textit{51} & \textit{178} & \textit{44} \\
   			
   			\midrule 
   	        \multirow{4}{*}{ Number of Pods}
   			& \multicolumn{1}{l}{Minimum} & \textit{144} & \textit{257} & \textit{142 }& \textit{142} \\
   			& \multicolumn{1}{l}{Maximum} & \textit{704} & \textit{831} & \textit{1058} & \textit{1038}\\
   			& \multicolumn{1}{l}{Mean}    & \textit{396.2} & \textit{423.9} & \textit{599.9} & \textit{597.2}\\
   			& \multicolumn{1}{l}{Standard deviation} & \textit{99.89} & \textit{106.61} & \textit{196.60} & \textit{198.17} \\    	
   		\end{tabular}
   	}
   \end{table}


\begin{figure}
    \includegraphics[width=16cm, height=9.5cm]{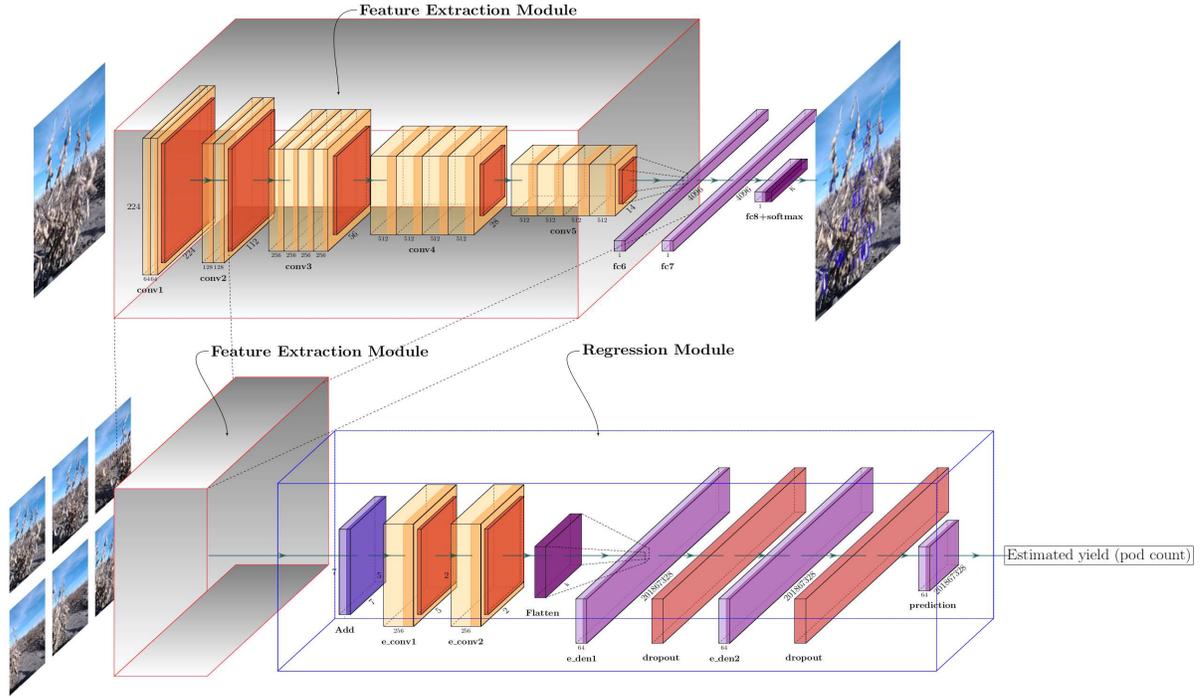}
	\caption{Yield (pod count) estimation model architecture consisting of a feature extraction module (FM) and a regression module (RM) diagram}
	\label{fig:ModelDiagram}
\end{figure}


\begin{figure} 
	\begin{subfigure}{0.48\textwidth}
		\includegraphics[width=7.5cm, height=8cm]{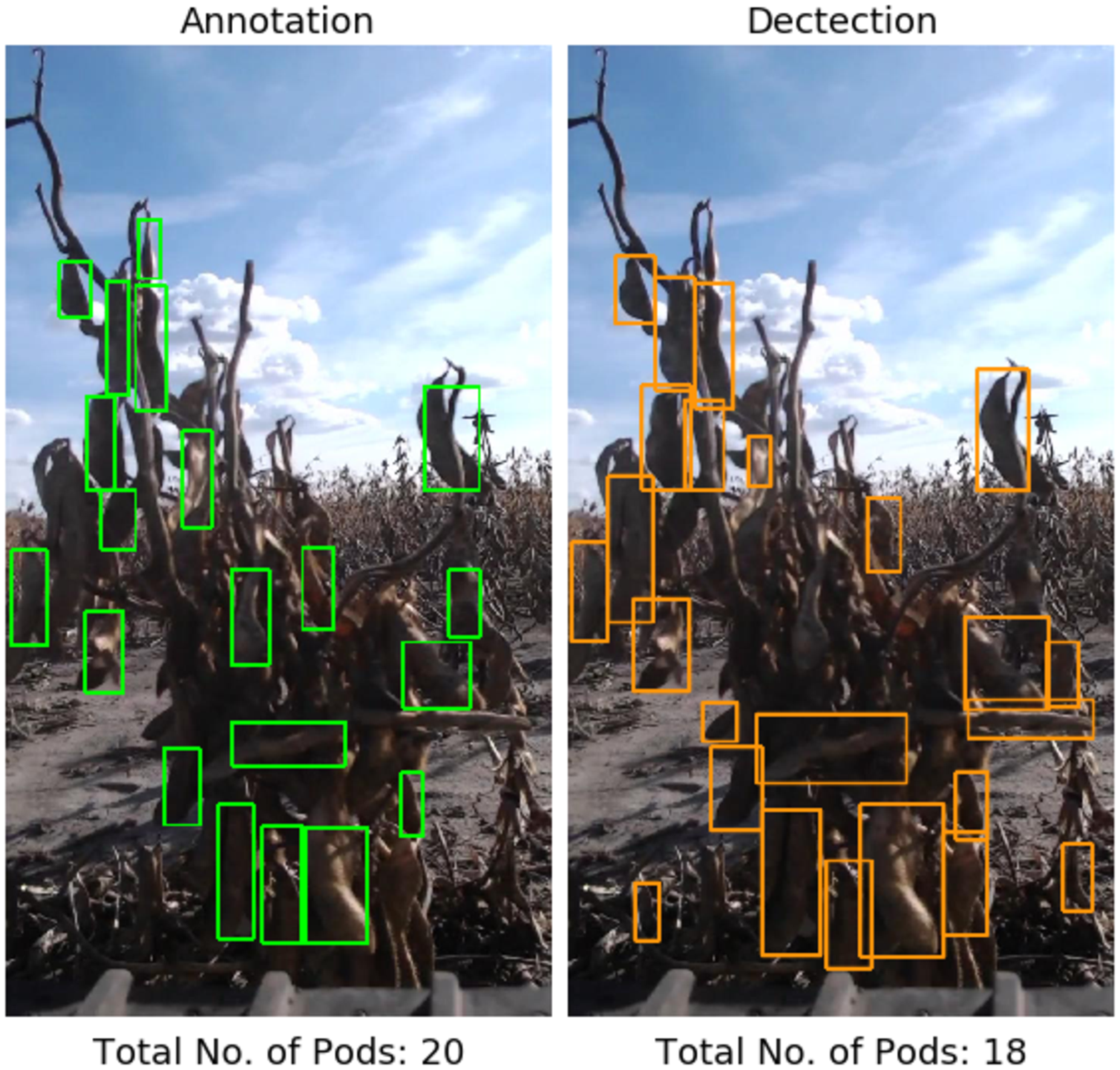}
	\end{subfigure}
	\begin{subfigure}{0.48\textwidth}
		\includegraphics[width=7.5cm, height=8cm]{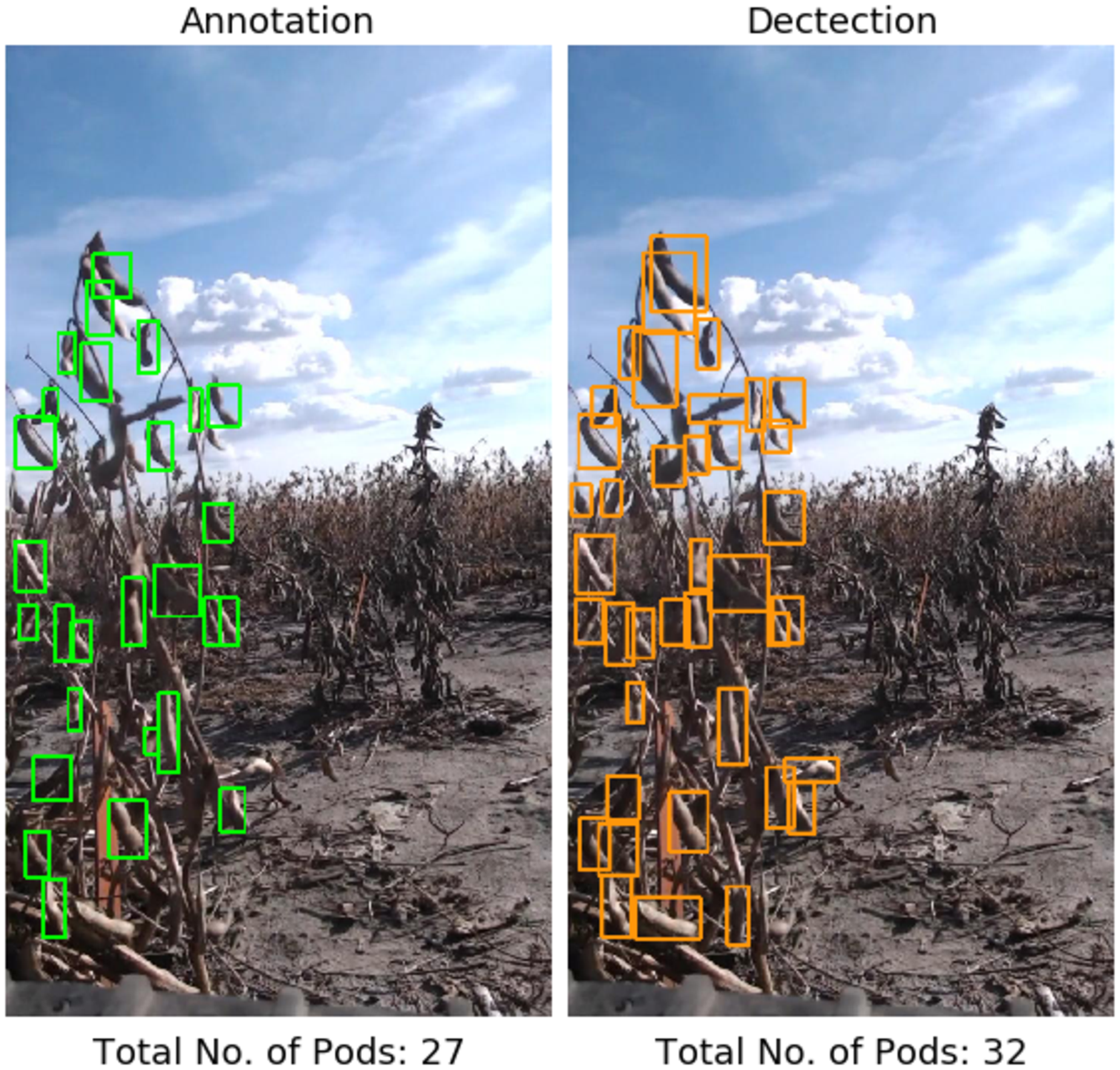}
	\end{subfigure}
	
	\begin{subfigure}{0.48\textwidth}
		\includegraphics[width=7.5cm, height=8cm]{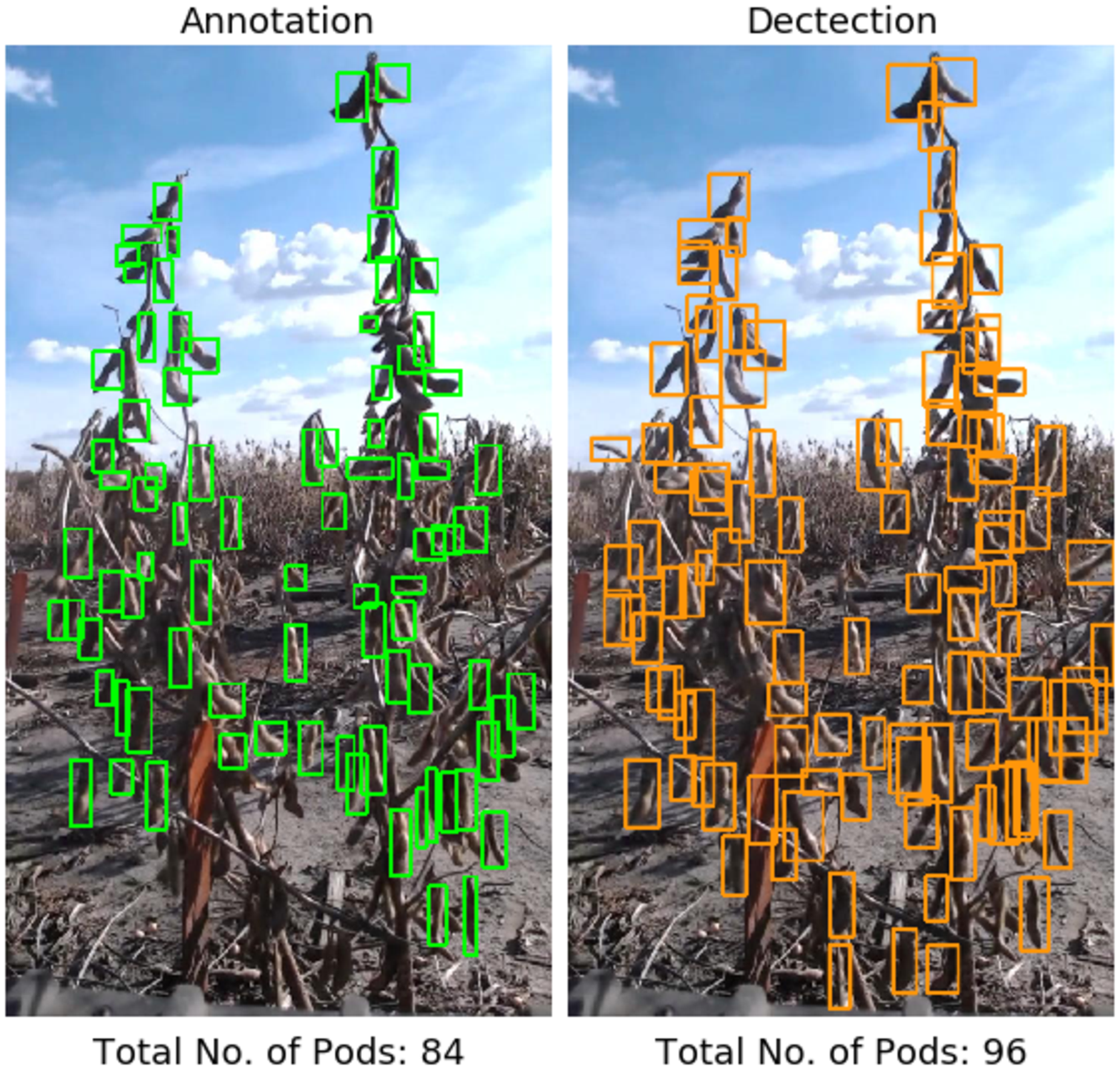}
	\end{subfigure}
	\begin{subfigure}{0.48\textwidth}
		\includegraphics[width=7.5cm, height=8cm]{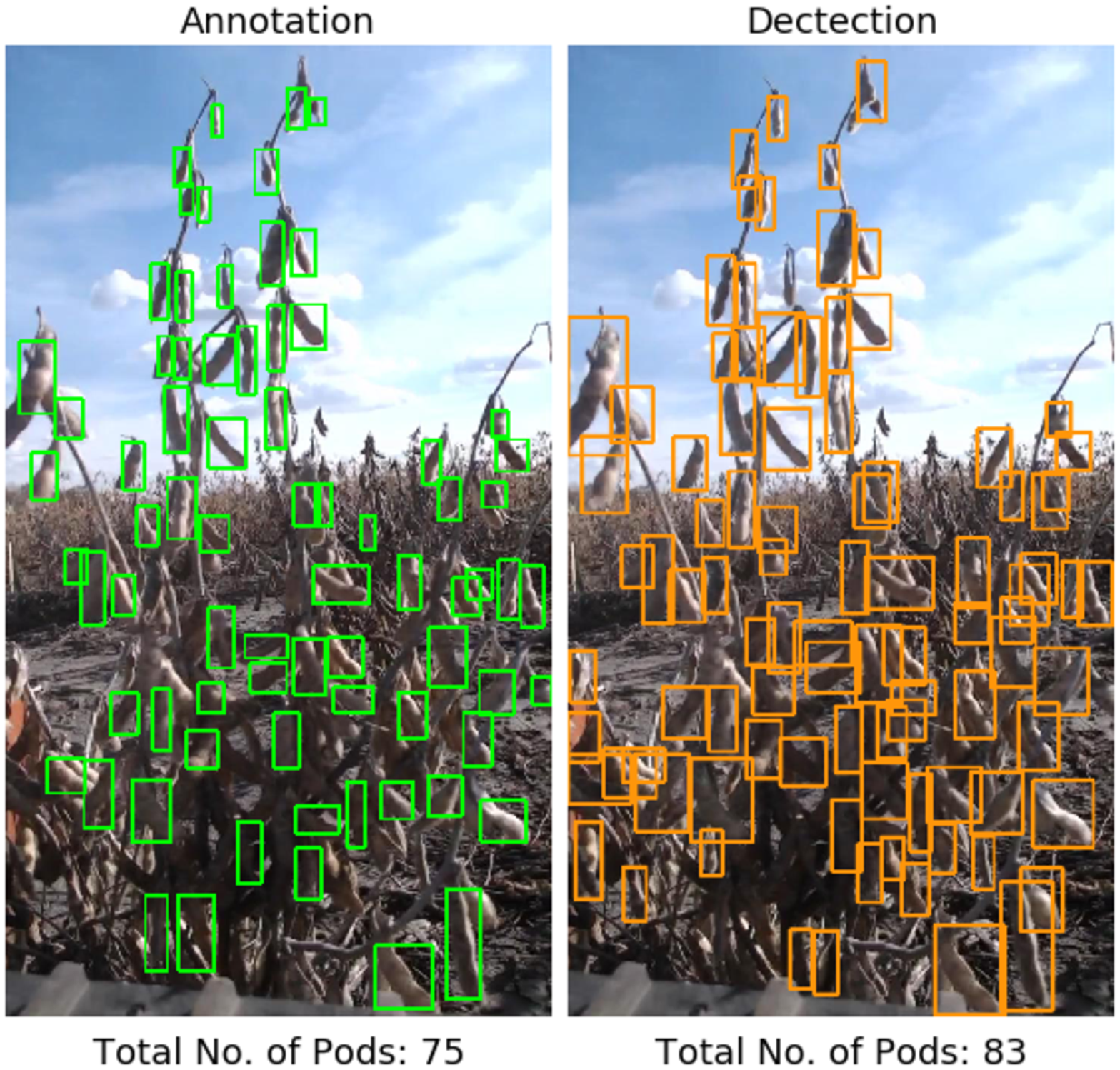}
	\end{subfigure}
	\caption{Plot samples with their respective annotated (left) and detected (right) pods.  Detection IoU threshold set to 0.5. Four different examples here show the data diversity in the \textit{in-field data set}}
	\label{fig:PodAnnotationDetection}
\end{figure}

\subsection{Machine learning framework for yield estimation}
\label{ssec:MachineLearningModel}
Soybean SY estimation through automated detection and counting of soybean pods is a challenging computer vision task. Complexity of this problem arises from various factors such as cluttered visual environment, occlusion of pods in a video frame and lighting variations. One possible approach to address these issues at least in part is to consider multiple video frames for a plot from different viewing angles. Therefore, we propose a deep learning-based multi-view image fusion framework that builds on a core model for pod detection and localization. In addition, to deploy this yield estimation framework on board a robotic platform, we need to detect and keep track of individual plots in real-time. In this regard, we also build a plot detection and tracking framework that can provide the necessary video frames for all plants in a specific plot to the yield estimation module. The proposed machine learning frameworks are described below.

\medskip

\subsubsection{Pod detection and yield estimation}
\label{sssec:PodDetection}
Our pod count (as measurement of seed yield) estimation model takes multiple RGB images of the same plot, with multiple plants in commercial planting density, from different viewing angles. The idea is that by having multiple images from different perspectives, the model could learn to overcome pods occlusion problems, mitigate possible image quality issues encountered during the automated images selection process from videos (i.e., sequence of frames), and data heterogeneity encountered in real-life soybean experiment, breeding and production fields. The model architecture is presented below:

\textbf{Model architecture and training process:} Our deep learning framework for multi-image fusion has two primary tasks - (i) pod detection on a soybean plot (with multiple plants) based on an individual image frame and (ii) estimating an overall pod count per plot based on multiple image frames. We choose a RetinaNet model architecture~\cite{lin2017focal} with a VGG19 backbone to execute the first task. During the first phase of training, we train the RetinaNet model to detect and isolate pods on soybean plants per plot as shown in Fig.~\ref{fig:PodAnnotationDetection}. For the \textit{control data set}, 99 images were randomly selected and annotated using a single class 'Pod' to train such a model. On the other hand, we use 513 randomly selected images from the \textit{in-field data set} for training a RetinaNet model for pod detection in a field setting.

After training and validation of a RetinaNet model, we focus on the next task of fusing information from multiple images to estimate the pod count for a soybean plot. The crux of the idea here is to leverage the features extracted by the pod detection model from the different images and map them to the overall pod count. To implement this, we use the lower 16 convolution layers of the VGG19 backbone of the trained RetinaNet model as the feature extractors (see~\cite{simonyan2014very} for the detailed structure of VGG19). We call this part of the model the feature extraction module (FM) as shown in Fig.~\ref{fig:ModelDiagram}. The features extracted by the FM layers from multiple images are concatenated and are used as inputs in a regression module (RM). The RM has three consecutive convolution layers, with their respective max-pooling and batch normalization layers, except for the last convolution layer.  These convolution layers are followed by a flatten layer, and three fully connected layers as shown in Fig.~\ref{fig:ModelDiagram}. The output of RM is the pod count for the plot consisting of multiple plants. We freeze the FM layers (taken from a well-trained RetinaNet model for pod detection) in order to train the RM layers. For the \textit{control data set}, we started with only the front view image of a plot as the input (to FM) and then added two side views for the multi-view version of the model. On the other hand, for the \textit{in-field data set}, input images come in pairs, taken from the opposite sides by the robotic platform. We experiment with one image from each side (for the single view model) as well as with three images from each side (for the multi-view model) of the plot. The training and test data distributions are already discussed in Section~\ref{ssec:ExperimentalSetup} and Table~\ref{tab:Dataset}. All model variations were trained and validated using a PC Workstation (OS: Ubuntu 18.04, CPU: Intel Xeon Silver 4108, GPU: Nvidia TITAN XP, RAM: 72 GB).

\subsubsection{Plant Detection and Tracking in Plots}
\label{sssec:PlantDetection}
In order to deploy the proposed yield estimation framework in an on-board, real-time fashion, we need to isolate the image frames corresponding to individual plots from the streaming video sequence collected by a camera on the robotic platform. However, in addition to isolating the image frames, if the plant area can be isolated in the image frame then that part of the image can be used for pod detection. This can help in reducing the negative effects of other plants (from other neighboring plots) in the background as well as other background clutters. Therefore, we first focus on detection and isolation of plants from a plot in video frames. 

\textbf{Model architecture and training process:} Similar to the pod detection model, we use a RetinaNet model with VGG19 backbone to detect and isolate the primary plot in an image frame. To train this model, we annotate about 1000 images from the \textit{in-field data set} with diverse background conditions as well as diverse shapes and sizes of soybean plants. We use 90\% of the annotated samples for training and rest for validation. 

Upon detection, we track the plants in a plot through the video frames with a unique ID tag such that the pod count estimation process can extract multiple frames for a specific plot and does not end up over-counting pods. There are two main aspects in a tracking algorithm. First, we detect and locate the targeted object in a frame, in our case, a soybean plot with multiple plants. Second, we decide whether the targeted object is present in subsequent frames. In our specific implementation, when the detector detects a plot, we save the information as central point of the bounding box. Each of the new central point is offered a unique ID and the location of those points are compared with the points in the subsequent frame, using Euclidean distance in a pair-wise manner. Based on the minimum distance, two central points (i.e.,  bounding  boxes) are assigned to the same plot. If a new plot appears, the central point of that plot is isolated, and a new unique ID is assigned. If the current plot disappears or does not get detected, the corresponding central point is saved as an existing point. The ID is removed if the corresponding central point does not appear for several frames (five frames in our implementation).

\medskip

%% file: Discussion.tex

In this section, we present the performance of our proposed framework in the context of soybean pod detection and pod count estimation. We also evaluate the usefulness of our pod count or yield estimation outcomes in breeding practices. Finally, for applicability of these outcomes in varied plots and images to ensure utilization in field breeding, we show some anecdotal performance of our plant detection and tracking framework.

\subsection{Pod detection performance}
\label{ssec:FeatureBlockValidation}
The pod detection model was validated using the Mean Average Precision (\textit{m}AP) metric, by setting the Intersection over Union (IoU) threshold at 0.55. Details of (\textit{m}AP) and IoU can be found in the supplementary material. The \textit{m}AP score for the \textit{control data set} and \textit{in-field data set} were 0.59 and 0.71, respectively. However, it is important to note that the \textit{control data set} had significantly lower pod annotations compared to its in-field counterpart. The rationale of having a smaller set of pod annotations in the \textit{control data set} was that even with such a small training set, the overall pod count estimation performance was acceptable (see \ref{ssec:EstimationBlockValitation}). Few anecdotal results for pod detection and isolation are presented in Fig.~\ref{fig:PodAnnotationDetection}.

    \begin{figure}[ht]
        \centering
        \begin{subfigure}{.49\textwidth}
            \centering
                \includegraphics[width =7.2cm, height = 5cm]{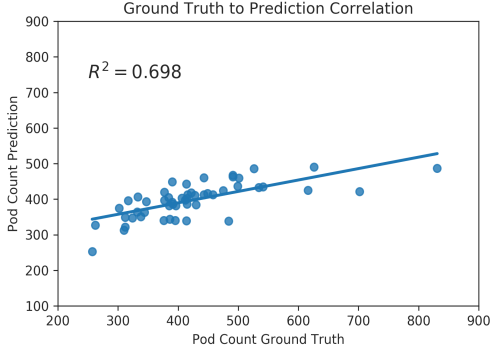}
                \caption{Correlation for the one image input model with \\ the \textit{control data set}}
                \label{fig:1-imgSet1}
        \end{subfigure}
        \begin{subfigure}{.49\textwidth}
                \centering
                \includegraphics[width =7.2cm, height = 5cm]{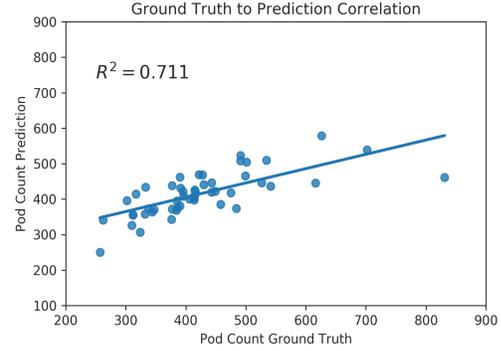}
                \caption{Correlation for the three images input model with \\ \textit{control data set}}
                \label{fig:3-imgSet1}
        \end{subfigure}
        \medskip
        
        \begin{subfigure}{.49\textwidth}
            \centering
                \includegraphics[width = 7.2cm, height = 5cm]{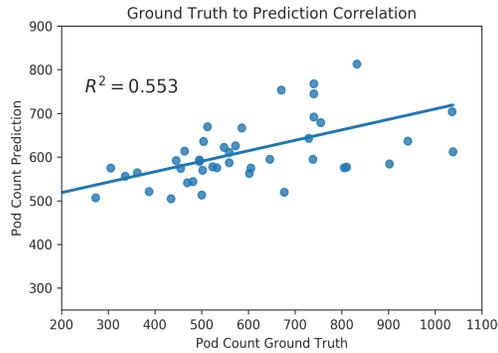}
                \caption{Correlation for the one image (per side) input \\ model with the \textit{in-field data set}}
                \label{fig:2-imgSet2}
        \end{subfigure}
        \begin{subfigure}{.49\textwidth}
                \centering
                \includegraphics[width = 7.2cm, height = 5cm]{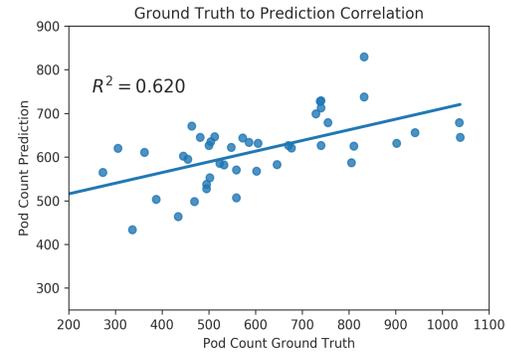}
                \caption{Correlation for the three images (per side) input \\ model with the \textit{in-field data set}}
                \label{fig:6-imgSet2}
        \end{subfigure}
        \caption{Correlations between ground truth and estimated pod counts using one and three images for the \textit{control} and \textit{in-field data sets}}
    \label{fig:ModelCorrelationPlots}
    
    \end{figure}

\subsection{Pod count estimation results}
\label{ssec:EstimationBlockValitation}
The correlation between the ground truth and the predicted pod counts for both the \textit{control data set} and the \textit{in-field data set} are provided in Fig. \ref{fig:ModelCorrelationPlots}. We observed that fusing multi-view images does help in improving the correlation between ground truth and prediction for both data sets. However, as expected, the performance is better for the \textit{control data set} (Fig 6 a,b), which can be attributed to the less occlusion and clean background with sharp color contrast with the foreground objects (plant and pods in this case). Interestingly, an improved pod count estimation performance for the \textit{control data set} was noted, despite using a feature extraction module that shows a lower \textit{m}AP value for pod detection and localization (due to smaller size of training data). Although, these moderate correlations may be acceptable for breeding purposes and applications, the predicted pod counts had narrower ranges compared to the ground truth pod count ranges. This can be attributed to the training data distribution shown in Fig.~\ref{fig:DatasetsDitribution} where most data points lie close to the mean value and we end up with unbalanced data sets with less representations from extreme pod count values.

\subsection{Genotype ranking performance}
\label{ssec:PhenotypeRanking}

\begin{figure}[ht]
        \includegraphics[width=16cm, height=7cm]{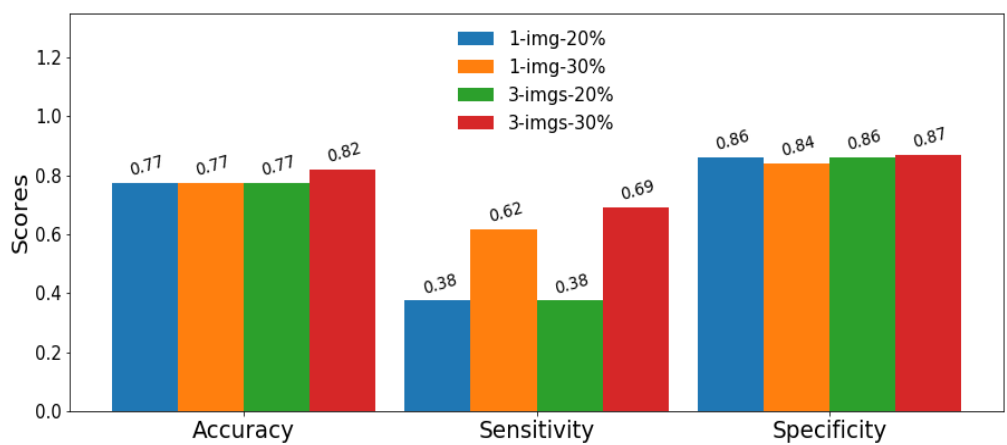} 
        \caption{Ranking scores for the one and three images per side models on the \textit{in-field} test data set}
        \label{fig:RankingScores}
    \end{figure}

While the correlation between pod count (i.e., proxy for yield estimation) prediction and the ground truth is an important metric for our proposed framework; from a breeding practice perspective, it is also important to make sure that the yield estimation framework is useful to downselect the top performing genotypes. For example, a 30\% selection cut-off means that only those plots that rank among the top 30\% (in terms of yield) are selected to advance to the next generation and subsequent testing (next season or year) in the breeding program. In this study, we use both 20\% and 30\% selection criteria to validate our framework, as these are reasonable downselect (i.e., culling) levels in a breeding program. Standard ML metrics such as accuracy, sensitivity and specificity are presented for evaluation [Fig.~\ref{fig:RankingScores} and Table~\ref{tab:RankingTestingResults}]. However, as our test data set sizes are rather small (51 test samples for \textit{control data set} and 44 test samples for \textit{in-field data set}), we also provide the actual numbers of True Positive, True Negative, False Positive and False Negative samples in Table~\ref{tab:RankingTestingResults}.

	\begin{table}[ht]
    \small
    
	\renewcommand{\arraystretch}{0.9}
	\caption{Model predicted ranking results for the top 20\% and 30\% plots from the \textit{in-field data set} using one and three images (per side) for pod counting}
	\label{tab:RankingTestingResults}
	\centering
	
	\makebox[\textwidth][c]{
		
	\begin{tabularx}{\linewidth}{l@{\hspace{0.25\tabcolsep}} 
	                            c @{\hspace{0.25\tabcolsep}} c |
	                            c @{\hspace{0.25\tabcolsep}} c | 
	                            c @{\hspace{0.25\tabcolsep}} c |
	                            c @{\hspace{0.25\tabcolsep}} c  
	                            }
		\toprule
		&
		 \multicolumn{2}{c|}{\textbf{1-img - Control Set}} &
		 \multicolumn{2}{c|}{\textbf{3-imgs - Control Set}} &
		 \multicolumn{2}{c|}{\textbf{1-img - In Field Set}} &
		 \multicolumn{2}{c}{\textbf{3-imgs - In Field Set}}
		\\
		\hline
		\midrule

            \multicolumn{1}{r}{\textbf{Ranking}} &
    		\multicolumn{1}{c }{\textit{Top 20\%}} & 
    		\multicolumn{1}{c|}{\textit{Top 30\%}} &
    		\multicolumn{1}{c }{\textit{Top 20\%}} & 
    		\multicolumn{1}{c|}{\textit{Top 30\%}} &
    		\multicolumn{1}{c }{\textit{Top 20\%}} & 
    		\multicolumn{1}{c|}{\textit{Top 30\%}} &
    		\multicolumn{1}{c }{\textit{Top 20\%}} & 
    		\multicolumn{1}{c}{\textit{Top 30\%}} 
    		\\
 
		\hline
		True Positive  & \textit{7} & \textit{12}  &  \textit{7} & \textit{10}  &  \textit{3} & \textit{8}  &  \textit{3 } & \textit{9}\\
		True Negative  & \textit{37} & \textit{33}  &  \textit{37} & \textit{31}  &  \textit{31} & \textit{26}  &  \textit{31} & \textit{27} \\
		False Positive & \textit{3} & \textit{3}  &  \textit{3} & \textit{5}  &  \textit{5} & \textit{5}  &  \textit{5} & \textit{4} \\
		False Negative & \textit{4} & \textit{3}  &  \textit{4} & \textit{5}  &  \textit{5} & \textit{5}  &  \textit{5} & \textit{4} \\
		
		\hline
		%
        Accuracy    & \textit{0.86} & \textit{0.88}  &  \textit{0.86} & \textit{0.80}  &  \textit{0.77} & \textit{0.77}  &  \textit{0.77} & \textit{0.82} \\
		Sensitivity & \textit{0.64} & \textit{0.80}  &  \textit{0.64} & \textit{0.67}  &  \textit{0.38} & \textit{0.62}  &  \textit{0.38} & \textit{0.69} \\
		Specificity & \textit{0.93} & \textit{0.92}  &  \textit{0.93} & \textit{0.86}  &  \textit{0.86} & \textit{0.84}  &  \textit{0.86} & \textit{0.87}\\

	\end{tabularx}
	}
\end{table}

From the results, it is clear that performance for the \textit{control data set} is slightly better compared to that for the \textit{in-field data set}, which conforms with our earlier correlation results, and is also true from the domain experience. However, we do not observe a significant improvement in performance with the usage of multi-view images as opposed to only single view images. For the \textit{in-field data set}, we noted that single view images from both sides of the plots are still needed [shown in Fig. 3(c-h)]. Overall, our results show that the proposed framework could be quite useful for selecting top performing genotypes especially when using a 30\% selection criteria compared to a 20\% selection criteria as evidenced with a higher sensitivity score. However, if a plant breeder is more interested in discarding the bottom performers, they can achieve reasonable success at 20\% selection level too due to high specificity and a lower sensitivity scores. This is particularly of importance in early stages of yield testing, where breeders are more concerned about "discarding" unworthy entries rather than "select" the top performers as the tests do not have sufficient statistical power to separate mean performance corresponding to the phenotypic and breeding values. With an improvement in test data size it is possible that model performance (i.e., sensitivity) may also improve enabling high confidence using more stringent downselection or culling levels. With small test data sets, it is difficult to draw strong conclusions in such a discrete classification setting (where, even one or two samples can change the overall statistics). Therefore, future work will focus on substantially increasing the test data set size to arrive at statistically more significant conclusions.

\subsection{Plot detection and tracking performance}
\label{ssec:plant_detection_results}
\begin{figure}[hbp]
    \centering
    \begin{center}
    \begin{tabular}{ p{3.35cm}  p{3.35cm} p{3.35cm} p{3.35cm}}
        \includegraphics[width=0.95\linewidth, height=5cm]{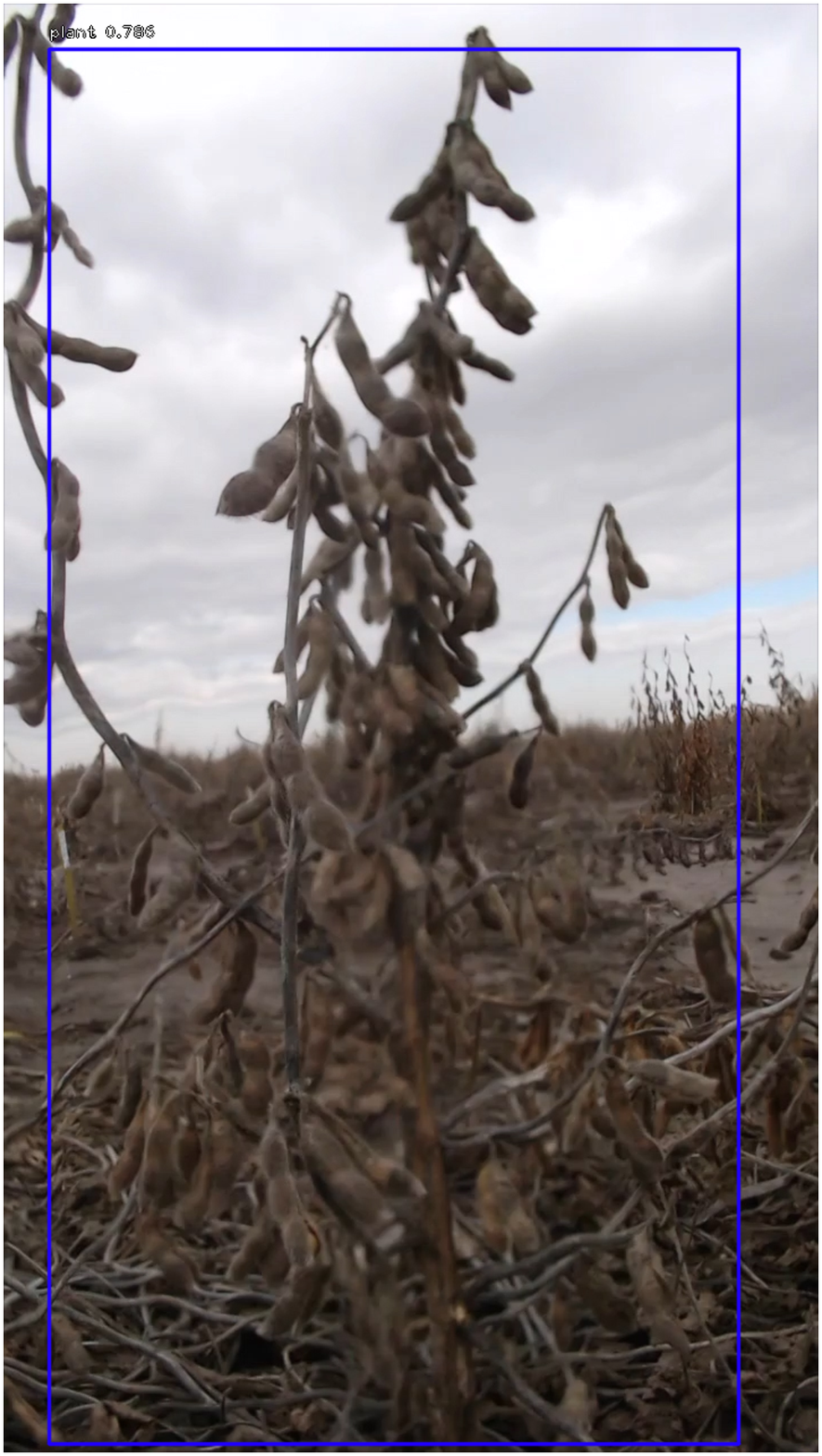}  &
        \includegraphics[width=0.95\linewidth, height=5cm]{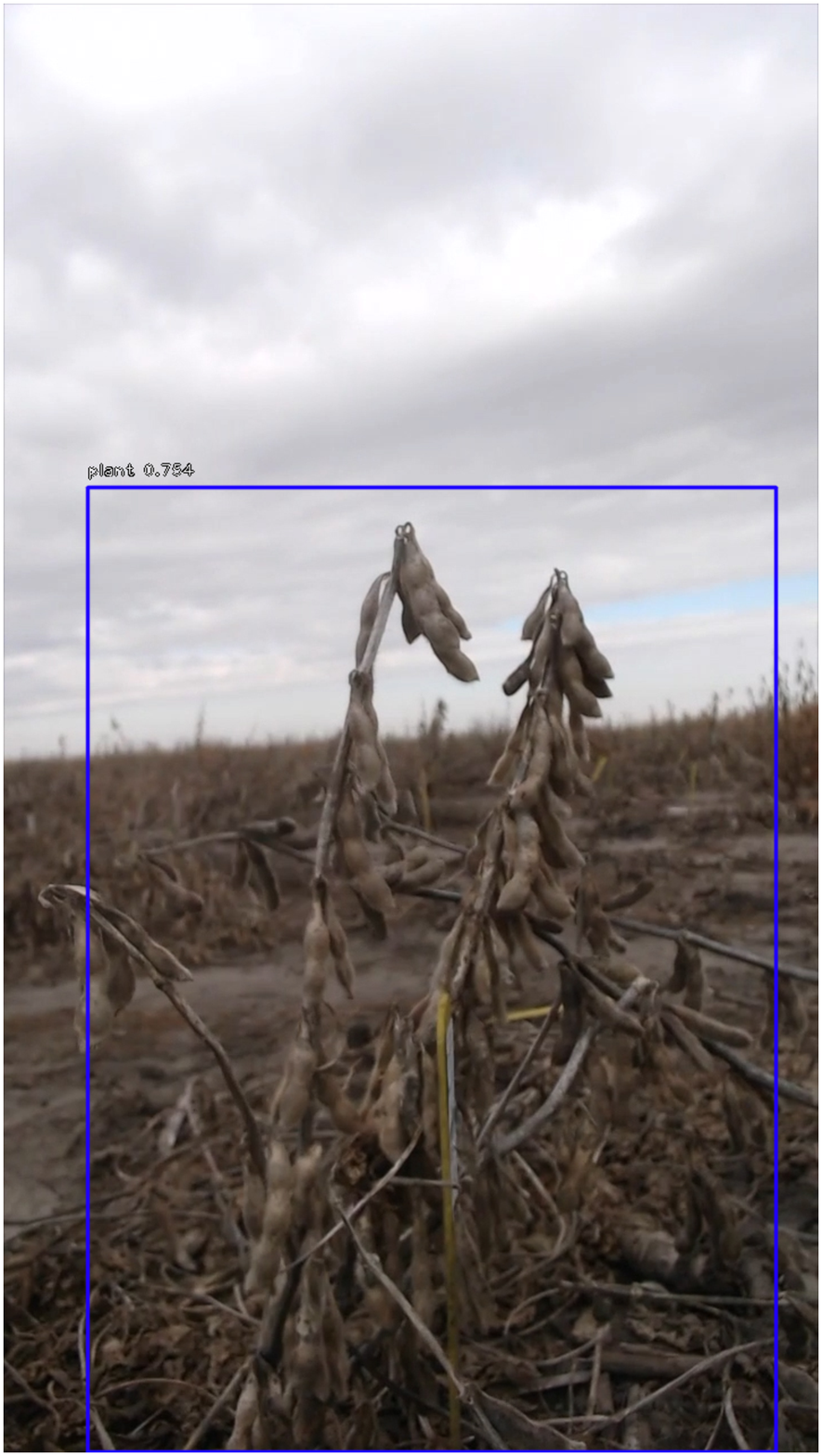}  &
        \includegraphics[width=0.95\linewidth, height=5cm]{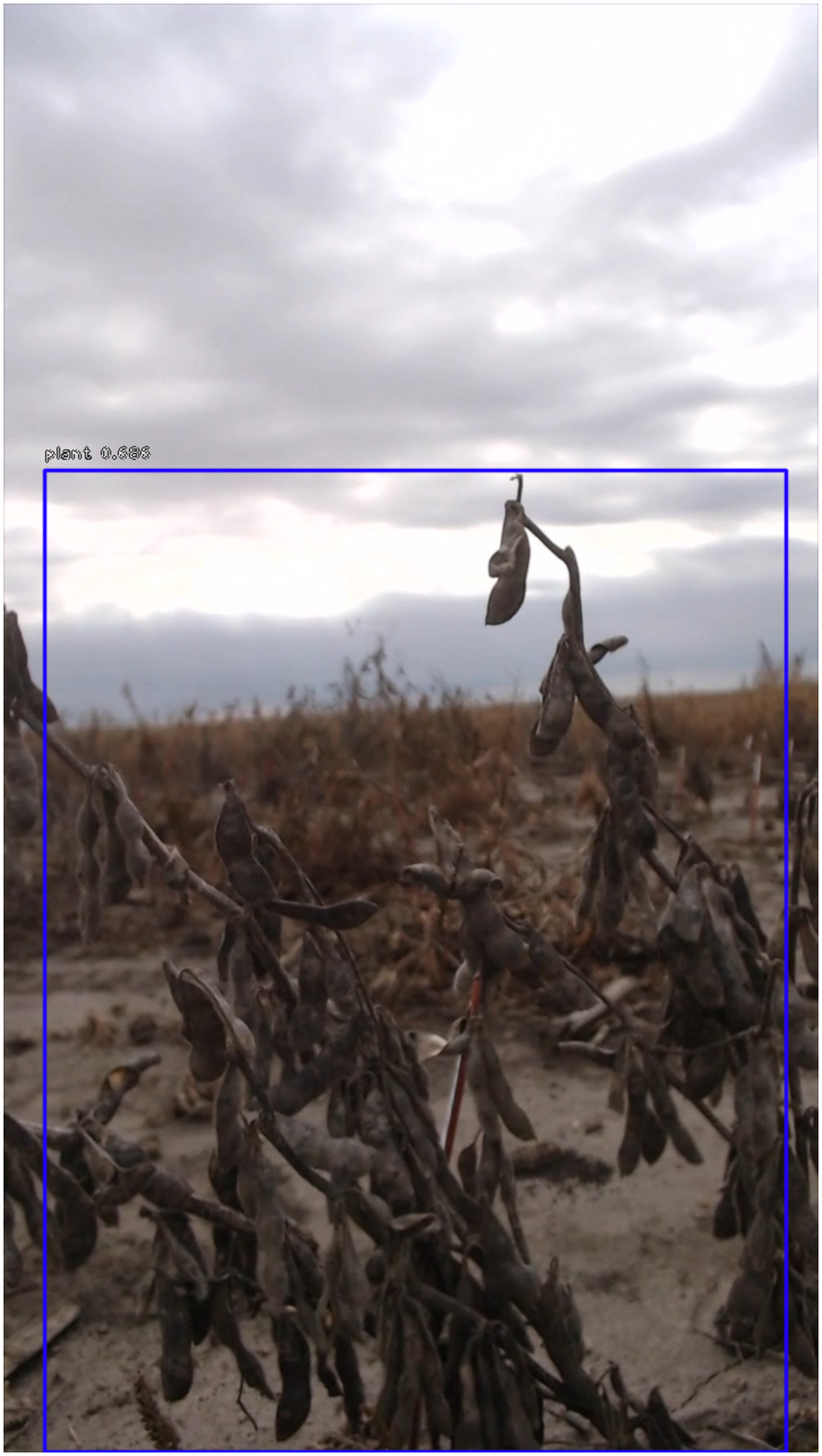}  &
        \includegraphics[width=0.95\linewidth, height=5cm]{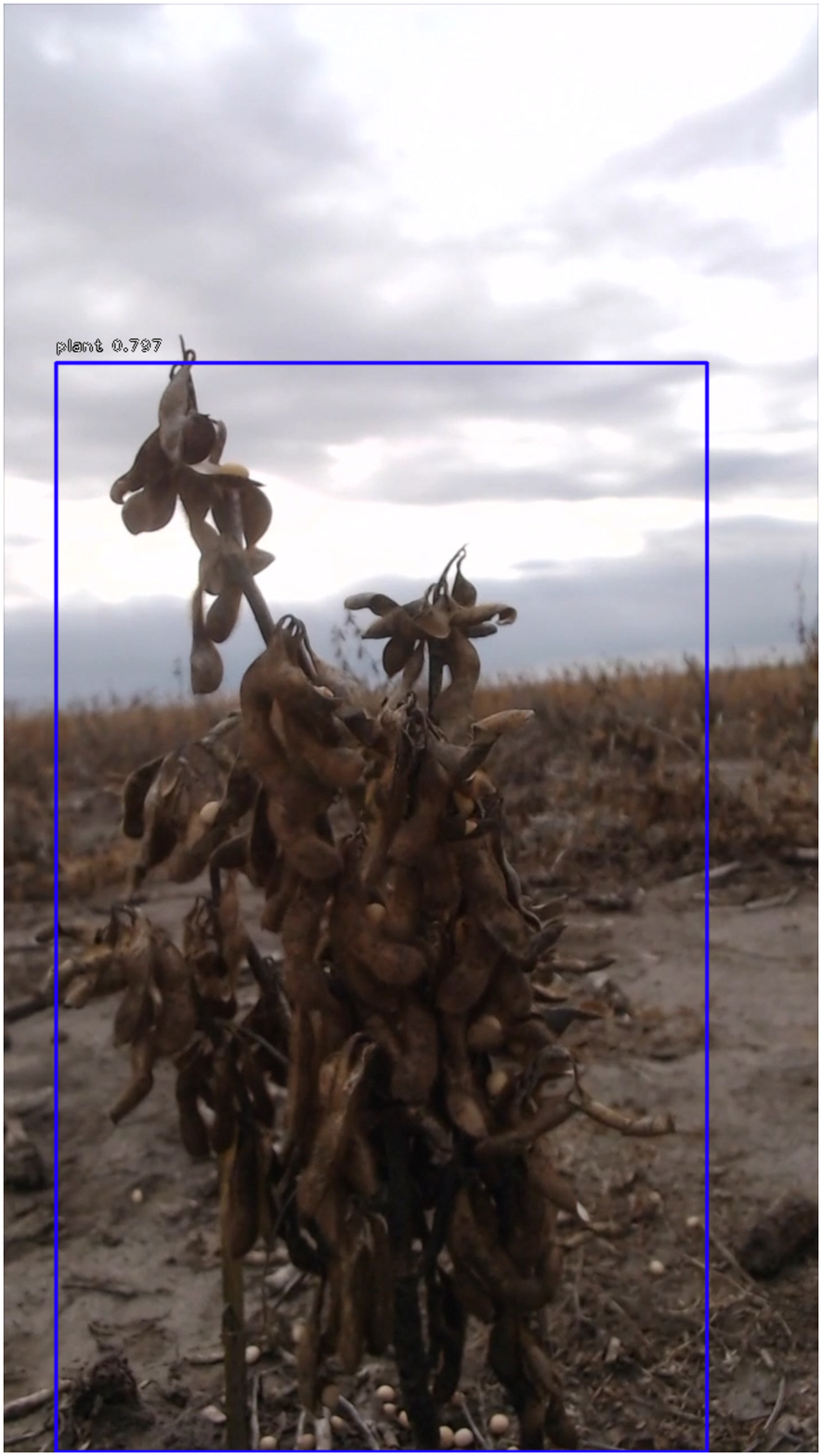}  \\
         \multicolumn{1}{c}{Plot 1} &
         \multicolumn{1}{c}{Plot 2} &
         \multicolumn{1}{c}{Plot 3} &
         \multicolumn{1}{c}{Plot 4} \\
        \multicolumn{4}{c}{(a) Samples of plots detected while the robot travels on the same row} \\
        \\
           
		\includegraphics[width=3.35cm, height=5cm]{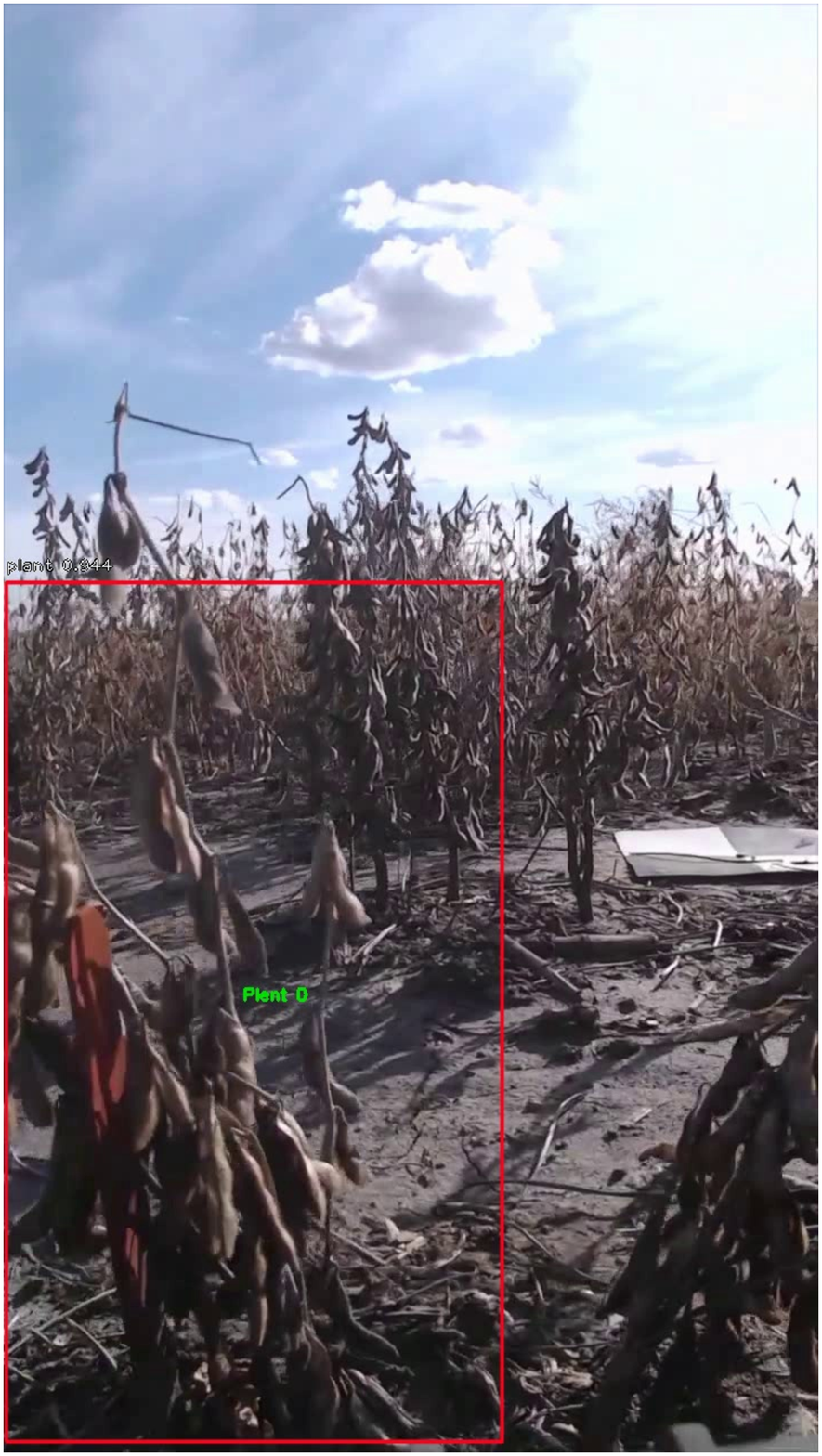} &
		\includegraphics[width=3.35cm, height=5cm]{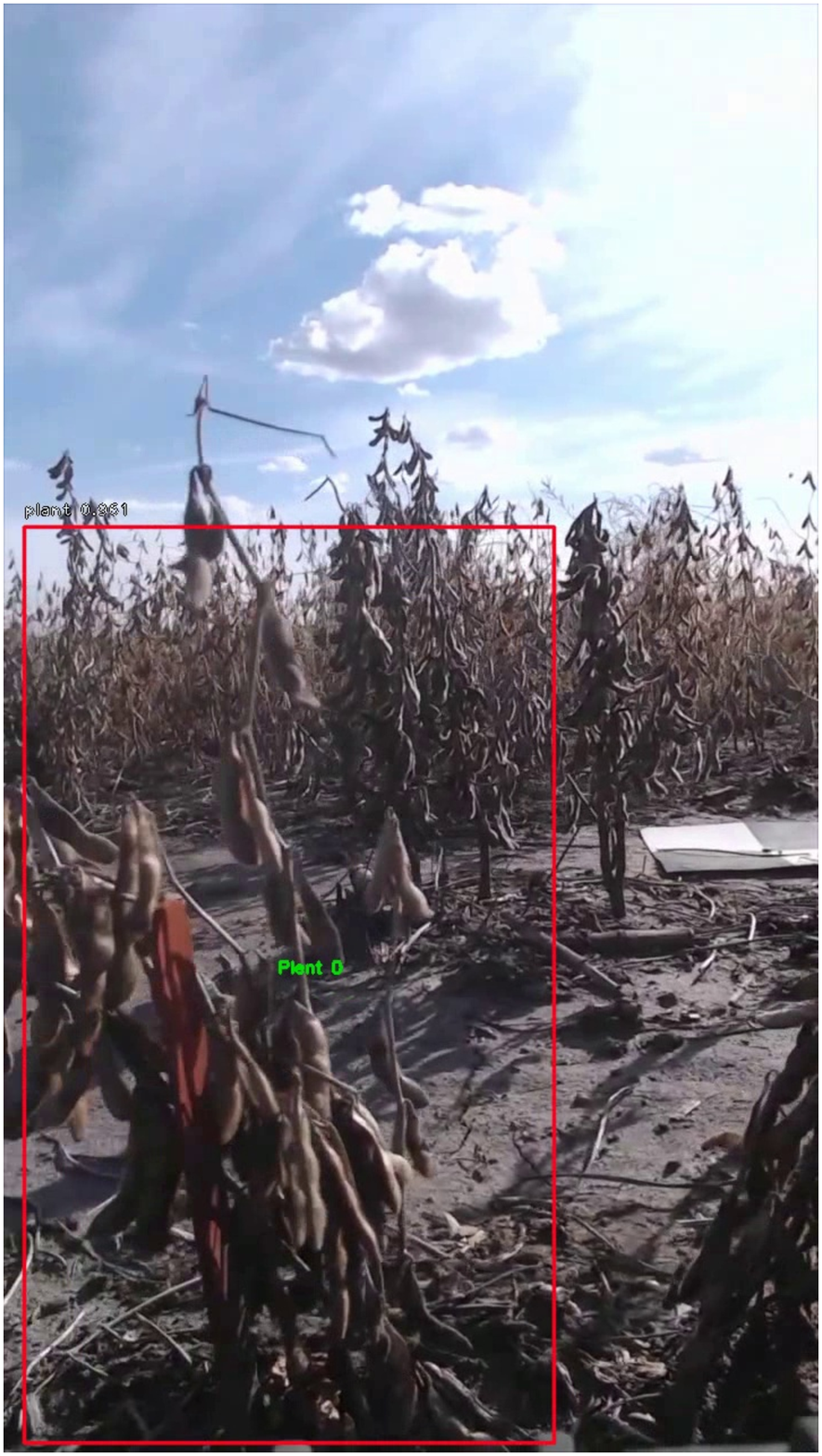} &
		\includegraphics[width=3.35cm, height=5cm]{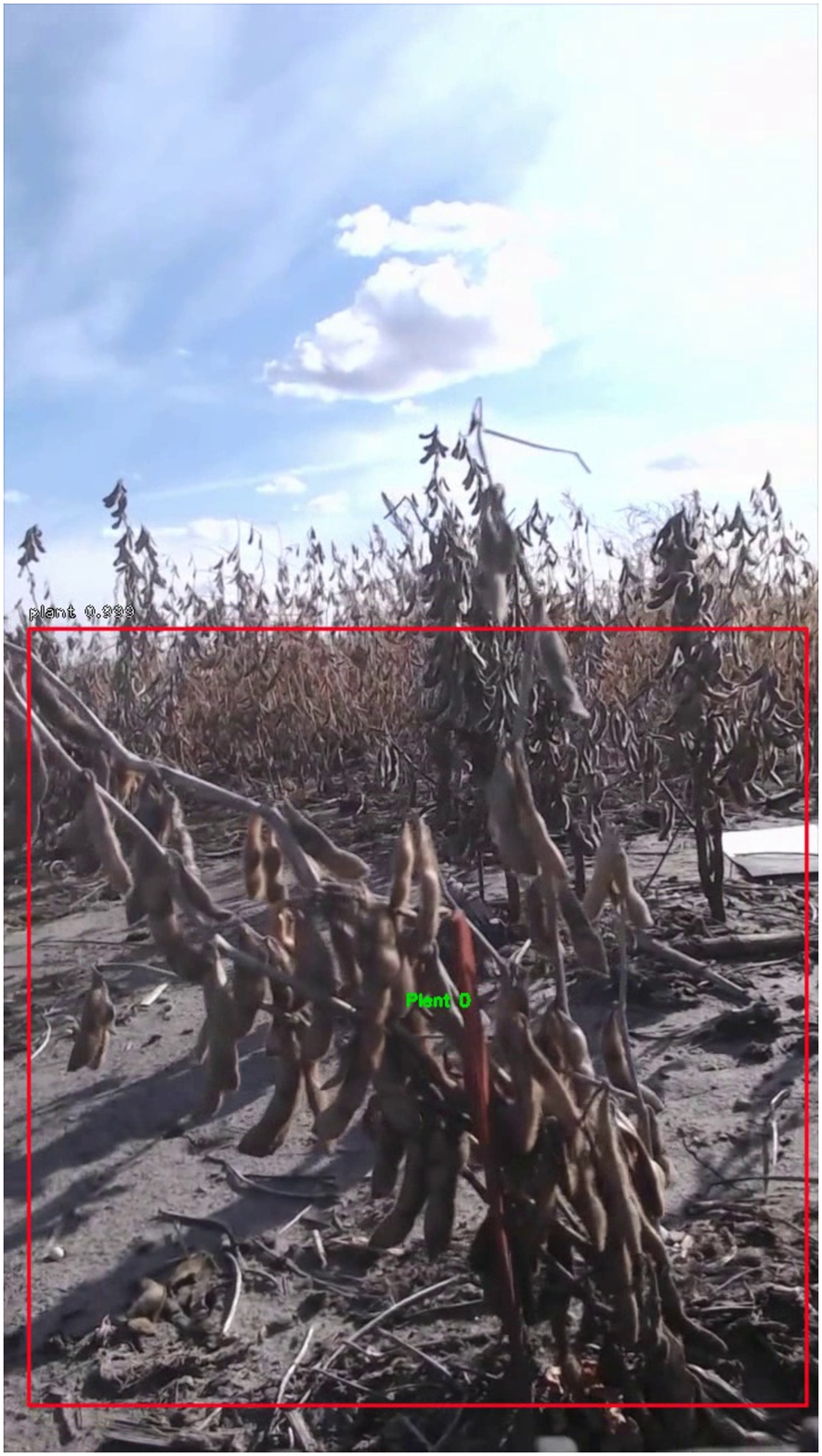} &
		\includegraphics[width=3.35cm, height=5cm]{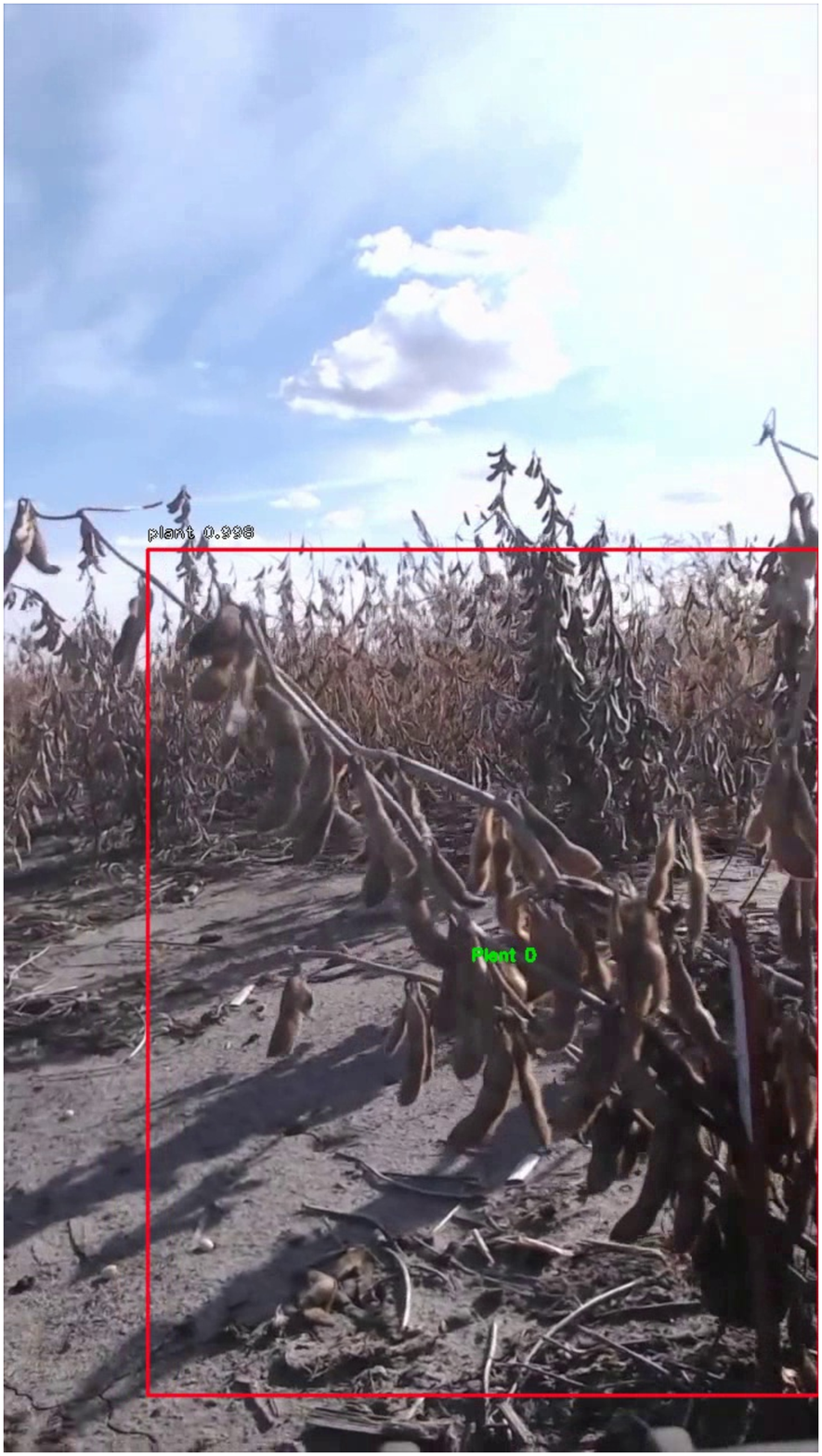} \\
		 \multicolumn{1}{c}{Frame No. 132} &
         \multicolumn{1}{c}{Frame No. 134} &
         \multicolumn{1}{c}{Frame No. 154} &
         \multicolumn{1}{c}{Frame No. 162} \\
        \multicolumn{4}{c}{(b) Sequences of images frames while detecting and tracking a single soybean plot from an input video} \\
        \\
	
    \end{tabular}
    \end{center}
	\caption{Samples of plot detection and tracking from recorded videos by the ground robot system.}
	\label{fig:PlantDetectionTracking}
\end{figure}

We provide few anecdotal results of our plot detection and tracking models in Fig.~\ref{fig:PlantDetectionTracking}. We report that plants of various sizes and shapes can be detected and sufficiently isolated despite a cluttered background with very low contrast. Although in practice we see that our plot detection and tracking system is mostly reliable, performance can suffer in low-light conditions as well as in severe occlusion scenarios, specifically due to large and lodged (non-upright) plants. Our future work will go beyond this anecdotal study to generate statistically significant quantitative results for a fully end-to-end on-board real-time soybean yield estimation system. However, we clearly show the feasibility of such a system in this paper.

%% file: conclusion.tex
In this paper, we propose a method capable of reducing overhead in yield testing trial. This was achieved with minimum human intervention by properly estimating yield, using pod count estimation, and ranking soybean genotypes for making breeding decisions. Our proposed method uses a deep learning framework that performs soybean pod detection and yield estimation using single or (fusing) multiple RGB images of a plant collected from a mobile motorized ground phenotyping unit. Our experiment showed relevant accurate results in a controlled outdoor environment; although, model performance was lower for the outdoor in-field image set, the results are still quite promising. We attribute the degradation of the model prediction to the lower quality of the data we had from the outdoor image set in comparison to the control one. The bottleneck of our experiments was not the ability to image plots, but the time and effort it took to manually count pods for every plot at a very high level of fidelity. One of the benefits of this type of yield estimation, is that it focuses on physically quantifying every pod in a plot, which has a direct correlation with the yield. We observed a correlation of 0.76 and 0.82 between manual pod count and seed yield for the  in-field and control sets respectively. An average of 2.0 and 2.2 seed/pod were noted for in-field and control set, respectively. 

The results of ML based pod counting method are now being integrated in our soybean breeding program at Iowa State University for breeding applications, while we continue to expand our research. First, we are exploring to improve the image gathering quality using an online feedback control system that interact with the robot to navigate the fields in an automated manner. Second, we are developing algorithms capable of matching video frames to accurate plant locations and determine the best frames to use for a plot. Both data sets were relatively small in size compared to many other use cases of deep learning frameworks. Therefore, increasing the data set size may show better results than what we were able to obtain. We note that active learning algorithms will also be useful to reduce the amount of labeling needed by deep learning models to achieve good predictive performance (\cite{nagasubramanian2020useful}).

Although, pod estimation is a proxy or surrogate trait for yield estimation, it is one of the important yield component of overall seed yield \cite{pandey1973path}. Most other yield prediction methodologies in plant science have focused on above canopy measurements such as reflectance measurements \cite{parmley2019development}, \cite{parmley2019machine} , and canopy coverage \cite{jubery2017deploying}. Future work should focus on the above problems, as well as moving integration of automated ways to obtaining yield component traits with indirect estimation of physiological traits and indices. These can be deployed at all field and controlled environment growing conditions, allowing for larger sampling sizes without the needs for labor intensive pod counting tasks and/or machine harvest of all plots. If high fidelity rankings can be achieved in full breeding plot tests, this methodology could help to greatly reduce the labor and time required for harvest operations in a breeding program in any give year. This will allow for less issues related to timely harvesting of plots, as well as faster decision making in a breeding program, with data available to a breeder sooner than would typically be available with traditional harvest methods. One important thing to consider however would be to appropriately analyze the data when it is in the ranking format instead of actual yield. In this case, it will be more difficult for a breeder to identify outliers, and perform spatial adjustments to a breeding trial when the plots are not set to yield, but rather to ranks. Although, with continual model advancements, actual yield prediction and not just rankings is not inconceivable. We are continuing to deploy these ML based methods for trait phenotpying including root nodule count - Soybean Nodule Acquisition Pipeline (SNAP) that quantifies nodule by combining RetinaNet and UNet deep learning architectures for object (i.e., nodule) detection and segmentation \cite{jubery2020using} as well as for microscopic nematode egg detection in cluttered images \cite{akintayo2018deep} . ML and more specifically DL methods continue to open previously inconceivable phenotyping doors for breeding and research application.

%% file: Supplementary.tex

The \textit{m}AP is calculated according to Eq.~\eqref{eq:mAP}, where TP, FP, TN, and FN denote the true positive, false positive, true negative and false negative, respectively.  The thresholds for in Eq.~\eqref{eq:mAP} was set to 0.5, this means that any predicted object is considered a TP if  its’ IoU with respect to the ground truth is greater than 0.5. 

	\begin{equation} 
	\label{eq:mAP}
	{\textit{m}AP} = \frac{1}{|threshold|} \sum_{t} \frac{TP(t)}{TP(t)+FP(t)+FN(t)}
	\end{equation}

Intersection over union (IoU) criteria is computed by Eq.~\eqref{eq:IoU}. 
	\begin{equation} 
	\label{eq:IoU}
	{IoU(A,B)} = \frac{A \cap B}{A \cup B} 
	\end{equation}

We calculated the model Specificity Eq.~\eqref{eq:Specificity} to measure its ability to correctly identify those plots that did not meet the Top ranking criterium and the Sensitivity Eq.~\eqref{eq:Sensitivity} measure that ability to identify those plots meet the selecting criterium correctly. At the same time, Accuracy Eq.~\eqref{eq:Acc_EQ} measured the closeness of the predictions to a specific value. 
    \begin{equation} 
    \label{eq:Specificity}
    \small
    	Specificity = \frac{TN}{TN + FP} 
    \end{equation}

    \begin{equation} 
    \label{eq:Sensitivity}
    \small
    	Recall(Sensitivity) = \frac{TP}{TP + FN} 
    \end{equation}
     \begin{equation} 
     \label{eq:Acc_EQ}
     \small
    	Accuracy = \frac{TP + TN} {TP + TN + FP + FN}
     \end{equation}

The correlation Eq.~\eqref{eq:CORR} was used as an indicator to measure the linear relationship between the model predictions and the ground truth values.
     \begin{equation} 
     \label{eq:CORR}
        \mathlarger{\mathlarger{\rho}}_{X,Y}~=~corr(X,Y)~=~\frac{cov(X,Y)}{\sigma_X\sigma_Y}~=~\frac{\mathbb{E}[(X-\mathbb{E}(X))(Y-\mathbb{E}(Y)]}{\sigma_X\sigma_Y}
     \end{equation}
     \\
\pagebreak

\textbf{Feature Module} architecture Summary\\
\small
\begin{tabular}{llc}
	\hline
	Layer (type)                 &Output Shape              &No. Param    \\
	\hline
	\hline
	input\_1 (InputLayer)         & (None, None, None, 3)    &  0 \\
	\hline         
	block1\_conv1 (Conv2D)        & (None, None, None, 64)   &  1792 \\      
	\hline 
	block1\_conv2 (Conv2D)        & (None, None, None, 64)   &  36,928  \\   
	\hline 
	block1\_pool (MaxPooling2D)   & (None, None, None, 64)   &  0       \\ 
	\hline 
	block2\_conv1 (Conv2D)        & (None, None, None, 128)  &  73,856   \\
	\hline 
	block2\_conv2 (Conv2D)        & (None, None, None, 128)  &  14,7584  \\
	\hline 
	block2\_pool (MaxPooling2D)   & (None, None, None, 128)  &  0     \\  
	\hline 
	block3\_conv1 (Conv2D)        & (None, None, None, 256)  &  29,5168  \\  
	\hline 
	block3\_conv2 (Conv2D)        & (None, None, None, 256)  &  590,080  \\
	\hline 
	block3\_conv3 (Conv2D)        & (None, None, None, 256)  &  590,080  \\
	\hline 
	block3\_conv4 (Conv2D)        & (None, None, None, 256)  &  590,080  \\
	\hline 
	block3\_pool (MaxPooling2D)   & (None, None, None, 256)  &  0       \\
	\hline 
	block4\_conv1 (Conv2D)        & (None, None, None, 512)  &  1,180,160  \\
	\hline 
	block4\_conv2 (Conv2D)        & (None, None, None, 512)  &  2,359,808  \\ 
	\hline 
	block4\_conv3 (Conv2D)        & (None, None, None, 512)  &  2,359,808   \\
	\hline 
	block4\_conv4 (Conv2D)        & (None, None, None, 512)  &  2,359,808   \\
	\hline 
	block4\_pool (MaxPooling2D)   & (None, None, None, 512)  &  0         \\
	\hline 
	block5\_conv1 (Conv2D)        & (None, None, None, 512)  &  2,359,808   \\
	\hline 
	block5\_conv2 (Conv2D)        & (None, None, None, 512)  &  2,359,808   \\
	\hline 
	block5\_conv3 (Conv2D)        & (None, None, None, 512)  &  2,359,808   \\
	\hline 
	block5\_conv4 (Conv2D)        & (None, None, None, 512)  &  2,359,808   \\
	\hline 
	block5\_pool (MaxPooling2D)   & (None, None, None, 512)  &  0         \\
	\hline 
	C5\_reduced (Conv2D)          & (None, None, None, 256)  &  131,328    \\
	\hline
\end{tabular}
\normalsize
\\
Total params: 20,155,712 \\
Trainable params: 20,155,712  \\

\textbf{Estimator Module} architecture summary,\\
\small
\begin{tabular}{llc}
	\hline
	Layer (type)                 &Output Shape              &No. Param    \\
	\hline
	\hline
	add\_1  (Add)                 &(None, 7, 7, 256)      &0 \\
	\hline
	e\_cov2 (Conv2D)              &(None, 5, 5, 256)     &590,080      \\ 
	\hline
	e\_cov2\_max (MaxPooling2D)    &(None, 2, 2, 256)     &0         \\ 
	\hline
	flatten\_1 (Flatten)          &(None, 1024)           &0         \\ 
	\hline
	e\_den1 (Dense)               &(None, 16)                &164 \\ 
	\hline
	e\_den2 (Dense)               &(None, 8)                &16      \\ 
	\hline
	e\_den3 (Dense)               &(None, 1)                 &9     \\
	\hline
\end{tabular}
\normalsize
\\
\\
Total params: 20,762,337 \\
Trainable params: 606,625 \\
Non-trainable params: 20,155,712\\